\documentclass[]{ustc_conference}
\usepackage{amssymb}
\usepackage[utf8]{inputenc}
\usepackage{booktabs}
\usepackage{multirow}
\usepackage{xcolor}
\usepackage{adjustbox}
\usepackage{array}
\usepackage[edges]{forest}
\usepackage{tikz}
\usepackage{amsfonts}
\usepackage[toc,page,header]{appendix}
\usepackage{bbm}
\usepackage{enumitem}
\usepackage{mathtools}
\usepackage{tabularx}
\renewcommand{\paragraph}[1]{\vspace{0.1em}\noindent\textbf{#1}}
\renewcommand{\titlefont}{\fontsize{17}{20}\selectfont}

\title{\protect\raisebox{-0.18em}{\protect\includegraphics[height=1.15em]{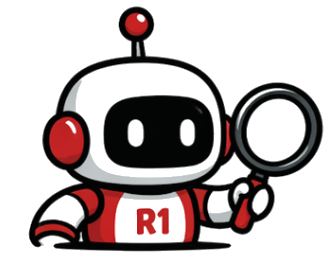}}\hspace{0.35em}Agent-R1: A Unified and Modular Framework for Agentic Reinforcement Learning}

\author{%
\parbox{\textwidth}{\centering
Mingyue Cheng, Shuo Yu, Daoyu Wang, Qingchuan Li, Xiaoyu Tao,  Jie Ouyang,    Yucong Luo,  Yitong Zhou,  Qi Liu, Enhong Chen
}}

\affiliation{%
\parbox{\textwidth}{\centering
State Key Laboratory of Cognitive Intelligence, University of Science and Technology of China
}}

\correspondence{\email{mycheng@ustc.edu.cn}}

\abstract{
Large language models (LLMs) have rapidly evolved from single-turn text generators into the foundation of increasingly capable agents. As these agents take on more complex reasoning, decision making, tool use, and long-horizon tasks, reinforcement learning (RL) is becoming increasingly important for shaping their behavior. This shift is especially visible in agentic RL, where models must interact with tools and environments across multiple rounds rather than produce a single standalone response. In this regime, the usual view of a trajectory as one ever-growing token sequence becomes increasingly inadequate: it makes context evolution rigid and creates representation mismatches between rollout and training. This paper presents \textbf{Agent-R1}, a unified and modular framework for agentic RL built around step-level trajectory representation, flexible context management, and layered interfaces for workflows, environments and optimization. The key idea is to treat each interaction step as the basic reinforcement-learning transition, while keeping the optimization layer flexible: once the interaction is modeled at the step level, the framework can support token-level credit assignment, step-level credit assignment, or other compatible designs. These design choices make the framework compatible with a range of optimization strategies rather than tying it to a single algorithm. Together, these components provide a principled, extensible, and reusable substrate for agentic RL.
}

\checkdata[Agent-R1]{\url{https://github.com/AgentR1/Agent-R1}}

\begin{document}

\maketitle

\begin{table}[t]
\centering
\caption{Representative open-source frameworks for agentic RL. Agent-R1 combines step-level MDP abstraction with flexible context management.}
\label{tab:granularity}
\footnotesize
\renewcommand{\arraystretch}{1.05}
\setlength{\tabcolsep}{10pt}
\begin{tabular}{@{}lll@{}}
\toprule
Framework & MDP abstraction & Context management \\
\midrule
veRL \cite{sheng2025hybridflow} & Token-level & No context management \\
slime \cite{slime2025} & Token-level & No context management \\
\mbox{Agent Lightning \cite{luo2025agentlightning}} & Step-level & Implicit context management \\
AReaL \cite{fu2025areal} & Not explicit & Implicit context management \\
rLLM \cite{rllm2026} & Not explicit & Implicit context management \\
\textbf{Agent-R1} \cite{agentr1repo} & \textbf{Step-level} & \textbf{Flexible context management} \\
\bottomrule
\end{tabular}
\end{table}

\section{Introduction}

Large language models (LLMs) \cite{achiam2023gpt,bai2023qwen} were initially developed as single-turn text generators, producing one response for one prompt. In this setting, they are mainly treated as conditional sequence models for tasks such as text continuation, question answering, summarization, and instruction following \cite{li2026llms}. As model capabilities improved, however, researchers began to build more complex systems around them. Through prompting, tool use, memory mechanisms, retrieval augmentation \cite{yu2025multi}, and environment feedback, LLMs evolved from standalone generators into the foundation of increasingly capable agents \cite{cheng2026comprehensive}. This shift has enabled systems that can search, plan, use tools \cite{yao2022react,schick2023toolformer,wang2023voyager}, and interact with environments over long horizons \cite{yao2022webshop,shridhar2020alfworld}. As these systems take on more ambitious goals, the challenge is no longer only to improve a single response, but to optimize the multi-turn decision process that drives sustained agent behavior.

Reinforcement learning (RL) is therefore becoming an increasingly important post-training paradigm for LLMs \cite{ouyang2022instructgpt,shao2024deepseekmath,wang2025ragen,zhang2025landscape}. Earlier recipes such as Reinforcement Learning from Human Feedback (RLHF) \cite{ouyang2022instructgpt} and Reinforcement Learning with Verifiable Rewards (RLVR) \cite{shao2024deepseekmath} were largely developed around single responses or short reasoning traces. In those settings, token-level generation remains a natural abstraction. Agentic RL introduces a different setting: the model must act across multiple rounds, interact with tools and environments, and adapt to external feedback over time. As a result, delayed and sparse rewards, long-horizon interaction, and long context all become first-class challenges. At the same time, the trajectory is no longer naturally viewed as a single growing response. Representing it as one ever-growing token sequence therefore becomes increasingly inadequate: it makes context evolution rigid and creates representation mismatches between rollout and training.

In this paper, Agent-R1 starts from a \textbf{step-level MDP} abstraction, in which each interaction round becomes a proper RL transition with an observation, an LLM action, and an environment update. This modeling choice makes the interaction step, rather than the token, the native unit for organizing agent behavior. 
Early agentic RL frameworks often stored trajectories as messages and later reconstructed them as text for training \cite{jin2025search}. While convenient, this creates a mismatch: rollout happens in token space, but training may rely on re-tokenized text. Since the token-text-token mapping is not reversible, this can introduce retokenization drift and break rollout-training consistency.
 Agent-R1 addresses this problem through a step-level trajectory representation in which each round preserves its observation and action boundary. It also addresses the rigidity of append-only context growth by allowing the environment to construct the next observation flexibly, rather than treating history as a flat sequence.

Table~\ref{tab:granularity} highlights the main distinction from representative agentic RL frameworks. Existing systems often make only one of these two design choices explicit: some retain token-level training abstractions, while others move closer to step-level interaction but leave context handling implicit. Agent-R1 emphasizes their combination within the same training substrate, making both the interaction unit and the context-construction rule explicit. This design not only preserves rollout structure while supporting task-dependent memory policies during training, but also lays a reusable foundation for future step-level credit-assignment research in the broader agentic RL community.

\section{Preliminaries}
\label{sec:preliminaries}

This section provides the background needed to understand why Agent-R1 is designed the way it is. We first review the current infrastructure landscape for LLM training and serving, then move from LLM training to LLM RL training, then from single-turn LLM RL to agentic RL, and finally summarize the concrete problems that arise once RL targets multi-turn agent behavior rather than single-turn text generation.

\subsection{Infrastructure for LLM Inference and Training}

The recent ecosystem of LLM infrastructure is already highly modular. At the inference and serving layer, engines such as vLLM \cite{kwon2023efficient} and SGLang \cite{sglang2024} provide efficient generation, batched decoding, and structured execution for LLMs. At the large-scale training layer, frameworks such as DeepSpeed \cite{deepspeedrepo}, PyTorch Fully Sharded Data Parallel (FSDP) \cite{zhao2023pytorch}, and Megatron-LM \cite{megatron-lm} support distributed optimization and parallel training at scale. Together, these systems have established a standard separation of concerns: high-throughput inference and large-scale optimization are often implemented as distinct layers rather than one monolithic stack.
This modular infrastructure is highly effective for conventional LLM development, but RL training must connect these two sides into one loop. Rollout depends on efficient inference engines, while optimization depends on scalable training frameworks. Once a model is embedded into an interactive workflow, the RL system must bridge them while also supporting tools, environment feedback, trajectory construction, and replay across many rounds. Like other agentic RL training frameworks, Agent-R1 is also designed to connect the inference side and the training side into a unified training loop. Agent-R1 is motivated by the observation that existing LLM infrastructure is necessary but not sufficient for this setting.

\subsection{From Supervised LLM Training to RL Training}

LLM RL training differs from standard supervised LLM training mainly in its execution loop. In standard supervised training, data are prepared in advance, batches are fed into the model, and optimization proceeds directly on the resulting losses. In LLM RL training, by contrast, optimization depends on rollout: the model must first generate responses, those responses must then be evaluated by rewards or preferences, and the resulting trajectories are sent back to the optimizer. This introduces a new dependency between inference and training that is largely absent from ordinary supervised learning.
Typical examples include RLHF \cite{ouyang2022instructgpt} and RLVR \cite{shao2024deepseekmath}, optimized with methods such as PPO \cite{schulman2017ppo} and GRPO \cite{shao2024deepseekmath}. Compared with standard LLM training, the pipeline must now support sampling, reward computation, and replay in addition to optimization itself. As a result, LLM RL training is not just a new loss function layered on top of LLM training; it is a different end-to-end loop that must coordinate rollout and optimization.

\subsection{From Single-Turn LLM RL to Agentic RL}
\begin{figure*}[t]
\centering
\includegraphics[width=\textwidth]{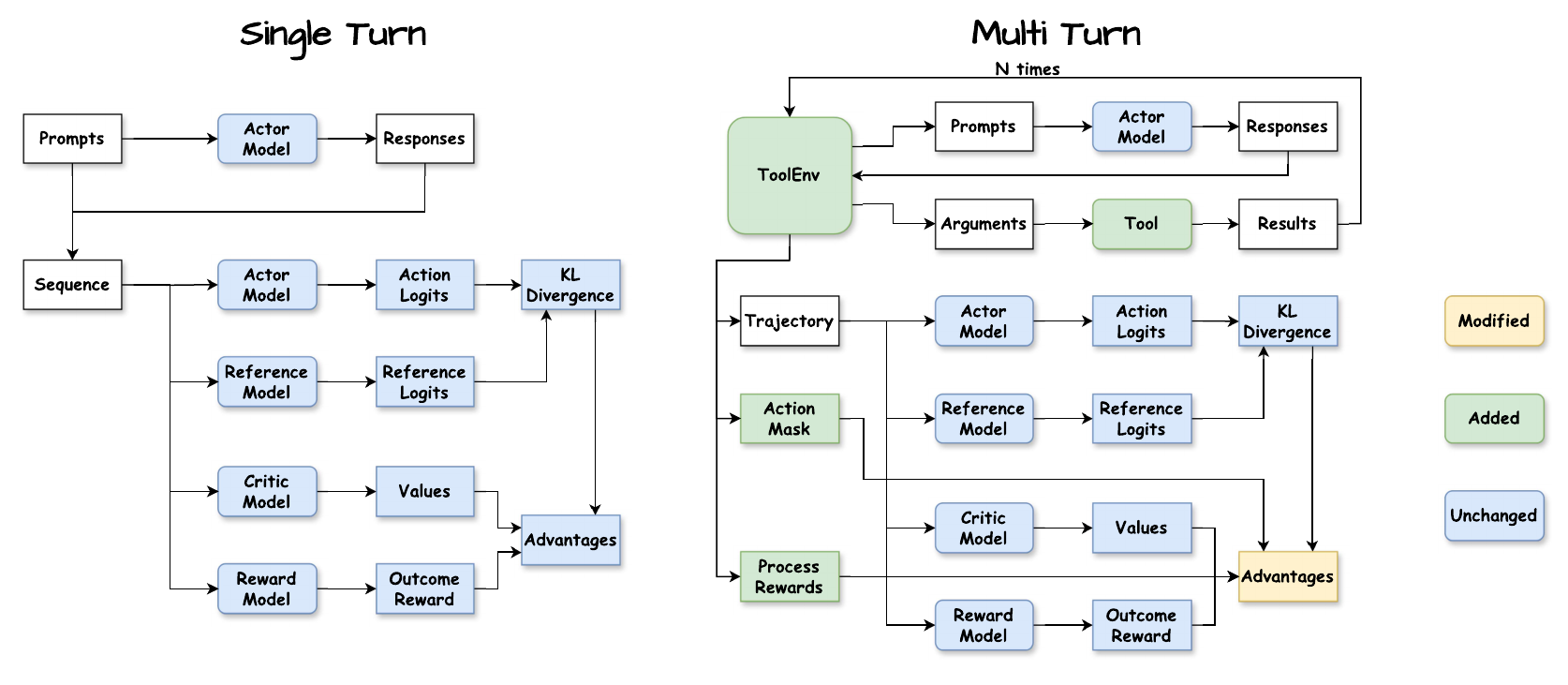}
\caption{Rollout in agentic RL. The agent repeatedly observes the current context, produces actions, and receives feedback from tools or environments, forming a multi-step trajectory rather than a single prompt-response pair.}
\label{fig:agentflow-generation}
\end{figure*}

\begin{figure*}[t]
\centering
\includegraphics[width=\textwidth]{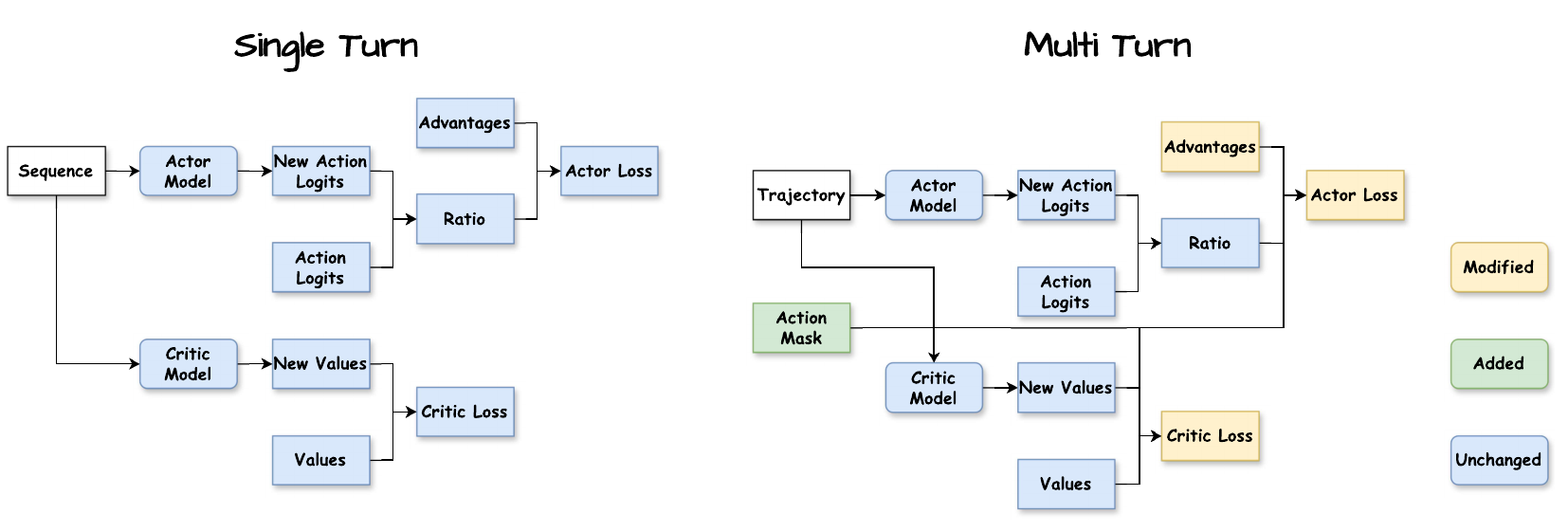}
\caption{Optimization in agentic RL. Interactive trajectories collected during rollout are replayed by the training loop and used to update the policy through reinforcement-learning objectives.}
\label{fig:agentflow-learning}
\end{figure*}

Agentic RL extends this change one step further. In single-turn or short-horizon LLM RL, rollout still mostly takes the form of one prompt and one response. In agentic RL, rollout becomes a multi-round interaction among the model, its tools, and the environment \cite{yao2022react,schick2023toolformer,yao2022webshop,shridhar2020alfworld,wang2025ragen}. As illustrated in Figures~\ref{fig:agentflow-generation} and~\ref{fig:agentflow-learning}, the framework must therefore connect rollout with trajectory replay and policy optimization, while handling environment feedback, multi-step trajectory construction, and interactive traces \cite{luo2025agentlightning,jiang2025verltool,fu2025areal}.
This shift changes what it means to preserve rollout faithfully. Rollout and optimization can no longer be treated as the same prompt-response abstraction; the framework must keep the causal structure of observation, action, feedback, and termination across multiple steps \cite{jiang2025verltool}. This requirement brings three practical consequences. First, trajectories need an explicit step-level representation rather than a flat generation sequence. Second, context construction must remain flexible, since different tasks may require selecting, reconstructing, or compressing prior interaction in different ways \cite{yao2022react,wang2025ragen}. Third, the system must connect LLM inference, tool execution, environment simulation, and optimization even though these components often operate at different timing and compute granularities \cite{yao2022webshop,shridhar2020alfworld,liu2023agentbench,luo2025agentlightning,fu2025areal,rllm2026}. These issues motivate the step-level formulation introduced next.

\subsection{From LLMs to Agents: An MDP Perspective}

The previous subsection described the practical shift from single-turn generation to multi-turn interaction in RL. The same shift can be formalized through the lens of Markov Decision Processes (MDPs). In the static single-turn setting, the state is typically the current textual context:
\begin{equation}
s_t = (\mathbf{w}_p, w_1, w_2, \ldots, w_t),
\end{equation}
where $\mathbf{w}_p$ is the initial prompt and $w_1,\ldots,w_t$ are the generated tokens. The action $a_t$ is the next token, and the transition is deterministic:
\begin{equation}
P(s_{t+1}\mid s_t,a_t)=
\begin{cases}
1, & \text{if } s_{t+1}=s_t\oplus a_t,\\
0, & \text{otherwise}.
\end{cases}
\end{equation}
This formulation is adequate for single-turn generation, but becomes restrictive once the model must interact with environments across multiple rounds.

For an LLM agent, the observable context must also capture prior interaction turns and environmental feedback. We therefore use an observation-centric step-level formulation:
\begin{equation}
o_t = (\mathbf{w}_p, \mathcal{T}_1, \mathcal{T}_2, \ldots, \mathcal{T}_k, \mathcal{T}_{k+1}^{\text{partial}}),
\end{equation}
where each $\mathcal{T}_i$ denotes a complete interaction turn and $\mathcal{T}_{k+1}^{\text{partial}}$ the ongoing partial turn. Under this view, Agent-R1 treats each interaction round, rather than each token, as the native action unit for training and optimization. Accordingly, a rollout is written as a sequence of structured step traces:
\begin{equation}
\tau = \{z_t\}_{t=0}^{T-1},
\qquad
z_t = (o_t, a_t, e_t, r_t, o_{t+1}),
\end{equation}
where $a_t$ may be a natural-language response, a structured tool invocation, or a mixed output containing both reasoning and external actions, and $e_t$ denotes the environment feedback returned after executing $a_t$.

At the interface level, the environment defines a step transition operator:
\begin{equation}
\mathcal{E}(o_t, a_t) = (o_{t+1}, r_t, d_t, e_t),
\end{equation}
where $d_t$ denotes termination and $e_t$ the environment feedback returned at step $t$. This step-level formulation naturally accommodates both terminal outcome rewards and intermediate process feedback, which is a key distinction between agentic RL and ordinary single-turn RL.

The same formalization also clarifies flexible context management. Rather than requiring the next observation to arise from simple append-only concatenation, Agent-R1 allows the environment to construct it explicitly:
\begin{equation}
o_{t+1} = \mathcal{C}(z_0, z_1, \ldots, z_t),
\end{equation}
where $\mathcal{C}$ is the context-construction rule applied to the structured interaction history. The resulting context may preserve or transform prior interaction history while remaining within a well-defined transition. This is why a step-level MDP provides a more natural foundation for multi-turn agent training.

\begin{figure*}[t]
\centering
\includegraphics[width=\textwidth]{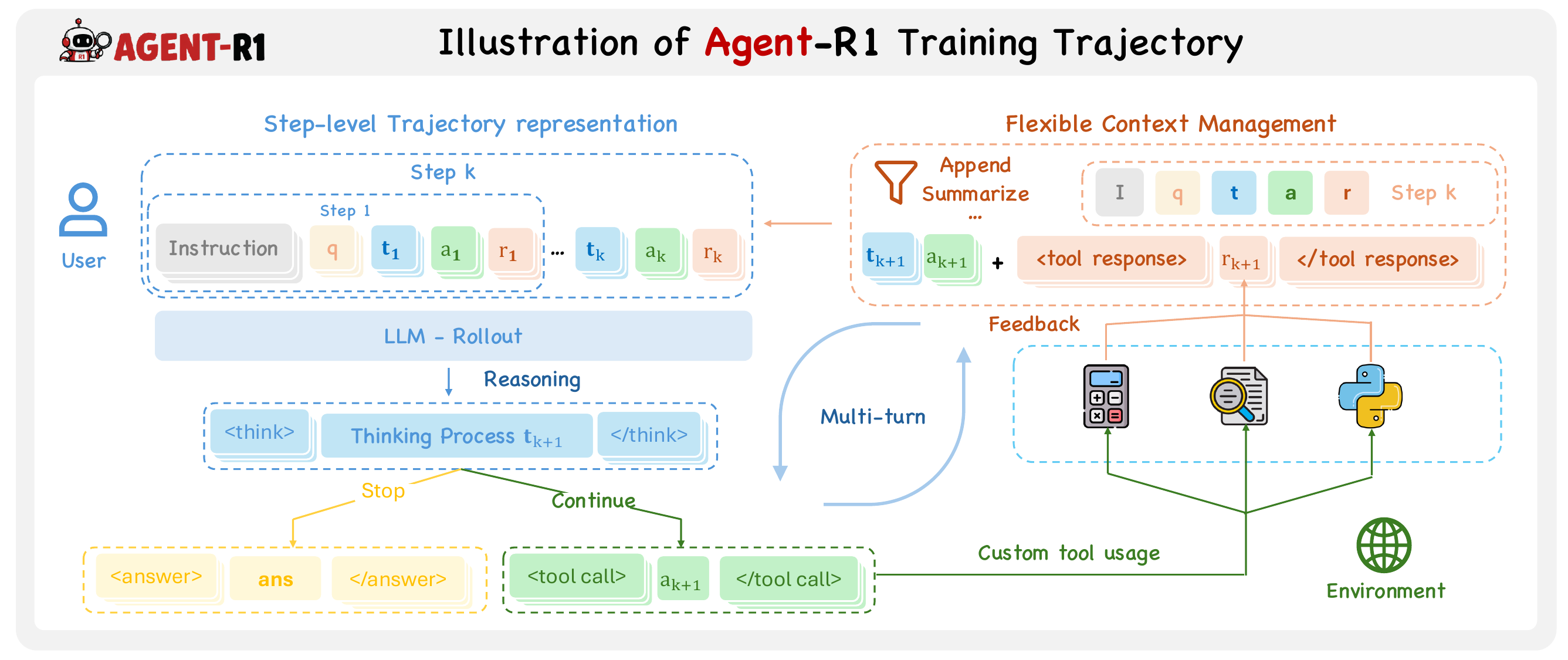}
\caption{Illustration of an Agent-R1 training trajectory. The left side shows step-level trajectory representation, where each interaction step is recorded as a structured unit rather than flattened into one long sequence. The right side shows flexible context management, where environment feedback can be appended, summarized, or otherwise reorganized before constructing the next-step context.}
\label{fig:mdp-formulation}
\end{figure*}

\section{Agent-R1}
\label{sec:mdp}

\subsection{Overview of Agent-R1}

The role of Agent-R1 is to connect and unify two sides that are often developed separately in practice. On one side are agentic RL algorithms, including optimization objectives, reward definitions, advantage estimation, and credit assignment. On the other side are infrastructure concerns, including workflow execution, rollout sampling, model serving, and large-scale optimization. Agent-R1 serves as the bridge between these two sides, so that algorithmic ideas can be instantiated on top of a common training substrate rather than being reimplemented together with the full execution stack each time.

Agent-R1 adopts interaction steps as the native training unit. Each step preserves its observation, action, feedback, and next observation, making step-level trajectory representation and flexible context management native to the training loop. At the interface level, the environment can be viewed as a step transition operator:
\begin{equation}
\mathcal{E}(o_t, a_t) = (o_{t+1}, r_t, d_t, e_t),
\end{equation}
where $o_t$ is the current observation, $a_t$ is the agent action for the current step, $o_{t+1}$ is the next observation, $r_t$ is the reward, $d_t$ is the termination flag, and $e_t$ denotes the environment feedback returned during the transition. Agent-R1 is designed so that workflows, tools, and optimization modules all communicate through this step-native interaction boundary.

To support optimization, rollout must preserve enough structure to distinguish complete actions, intermediate feedback, and final outcomes. A rollout can therefore be written as a sequence of step traces:
\begin{equation}
\tau = \{z_t\}_{t=0}^{T-1},
\qquad
z_t = (o_t, a_t, e_t, r_t, o_{t+1}),
\end{equation}
where $e_t$ denotes the environment-side feedback generated after executing action $a_t$. This representation allows the learning side to distinguish complete agent actions from environment, attach both outcome rewards and process rewards to the appropriate steps, and keep replay faithful to the interaction trajectory.

For optimization, one useful abstraction is the action mask over generated tokens within each step. If the token sequence emitted at step $t$ is written as $a_t = (y_{t,1}, \ldots, y_{t,L_t})$, then a mask is defined as:
\begin{equation}
m_{t,j} =
\begin{cases}
1, & \text{if } y_{t,j} \text{ belongs to the agent action at step } t,\\
0, & \text{otherwise}
\end{cases}
\end{equation}
This mask selects the tokens that should receive policy-gradient updates, while leaving prompt tokens and environment-side content outside the policy loss. A generic masked policy objective can then be written as:
\begin{equation}
\mathcal{L}_{\mathrm{policy}}
=
- \sum_{t=0}^{T-1}\sum_{j=1}^{L_t}
m_{t,j}\,\hat{A}_{t,j}\,
\log \pi_{\theta}(y_{t,j}\mid o_t, y_{t,<j}),
\end{equation}
where the credit term $\hat{A}_{t,j}$ may be instantiated either at token level or by broadcasting a step-level signal across the action tokens of step $t$. In this way, Agent-R1 standardizes the trajectory substrate used by optimization while remaining compatible with PPO-style, GRPO-style, and other RL objectives.

\begin{figure*}[t]
\centering
\includegraphics[width=\textwidth]{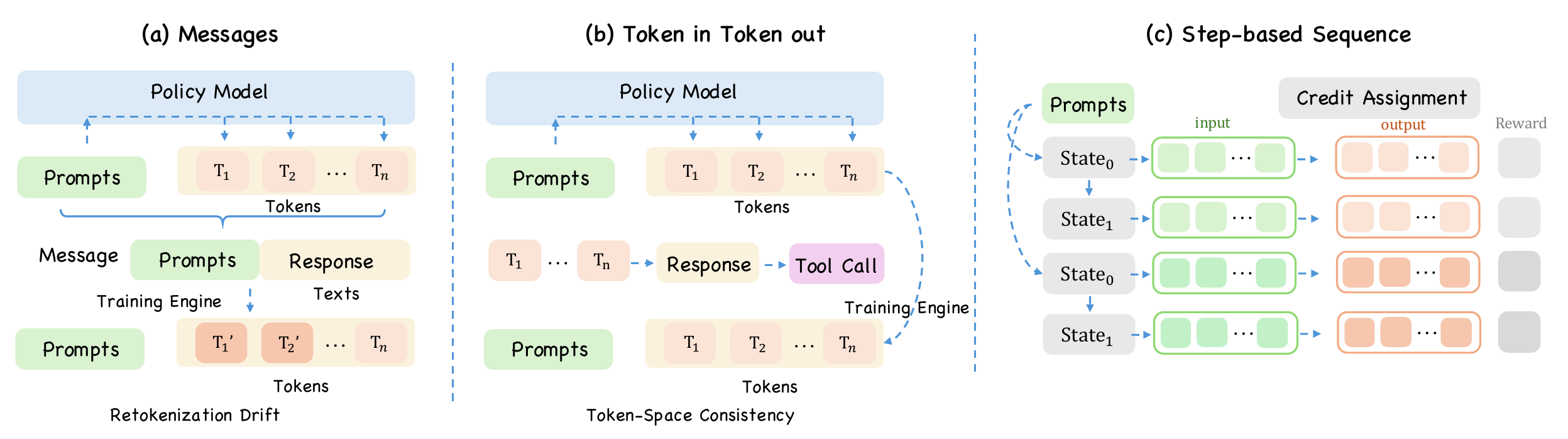}
\caption{Trajectory representation evolves from message-based traces to token-consistent records and finally to structured step-level traces. This progression is important not only for replay correctness, but also for flexible context management in multi-turn agent training.}
\label{fig:trajectory-representation}
\end{figure*}

\subsection{Step-level Trajectory Representation}

A central design choice in Agent-R1 is how multi-turn trajectories are represented for replay and optimization. In existing agent training pipelines, trajectory representations often fall into two common forms: message traces and flat token sequences. Message traces are convenient for workflow construction and debugging, but they do not preserve the exact token sequence generated during rollout. If $\mathrm{Tok}(\cdot)$ denotes tokenization and $\mathrm{Text}(\cdot)$ denotes text reconstruction, the replayed sequence is typically:
\begin{equation}
\tilde{x}_{1:N'} = \mathrm{Tok}\!\big(\mathrm{Text}(\mathcal{M}(\tau))\big),
\end{equation}
whereas the original rollout is generated on:
\begin{equation}
x_{1:N} = (x_1, x_2, \ldots, x_N).
\end{equation}
As discussed earlier, these two sequences need not be identical \cite{jiang2025verltool}. Such mismatch may shift action boundaries, alter the effective action mask, and distort the log-probabilities used during optimization. Flat token-space storage avoids this retokenization problem by preserving the exact rollout tokens, but it still treats the interaction as one append-only sequence and leaves step boundaries implicit.

Agent-R1 addresses these limitations by adopting step-level traces as the native trajectory abstraction. As introduced above, each rollout is stored as a sequence of structured step records, so that replay remains faithful to the original interaction while the step boundary stays explicit. By making the step boundary explicit, Agent-R1 can identify not only which tokens were generated, but also which token span corresponds to one complete agent action, what feedback it triggered, and how that feedback changed the next observation. If the action at step $t$ consists of tokens $(y_{t,1}, \ldots, y_{t,L_t})$, then rewards, masks, and credit signals can be aligned either with these internal action tokens or with the step $z_t$ as a whole.

The practical benefit is that step-level traces provide a common substrate for multiple optimization views. Token-level objectives can still be applied over the action tokens within each step, while step-level rewards or process supervision can be attached directly to the corresponding transition. In this way, step-level replay is not a derived convenience, but the native representation used by Agent-R1 for multi-turn agent RL.
\begin{figure*}[t]
\centering
\includegraphics[width=\textwidth]{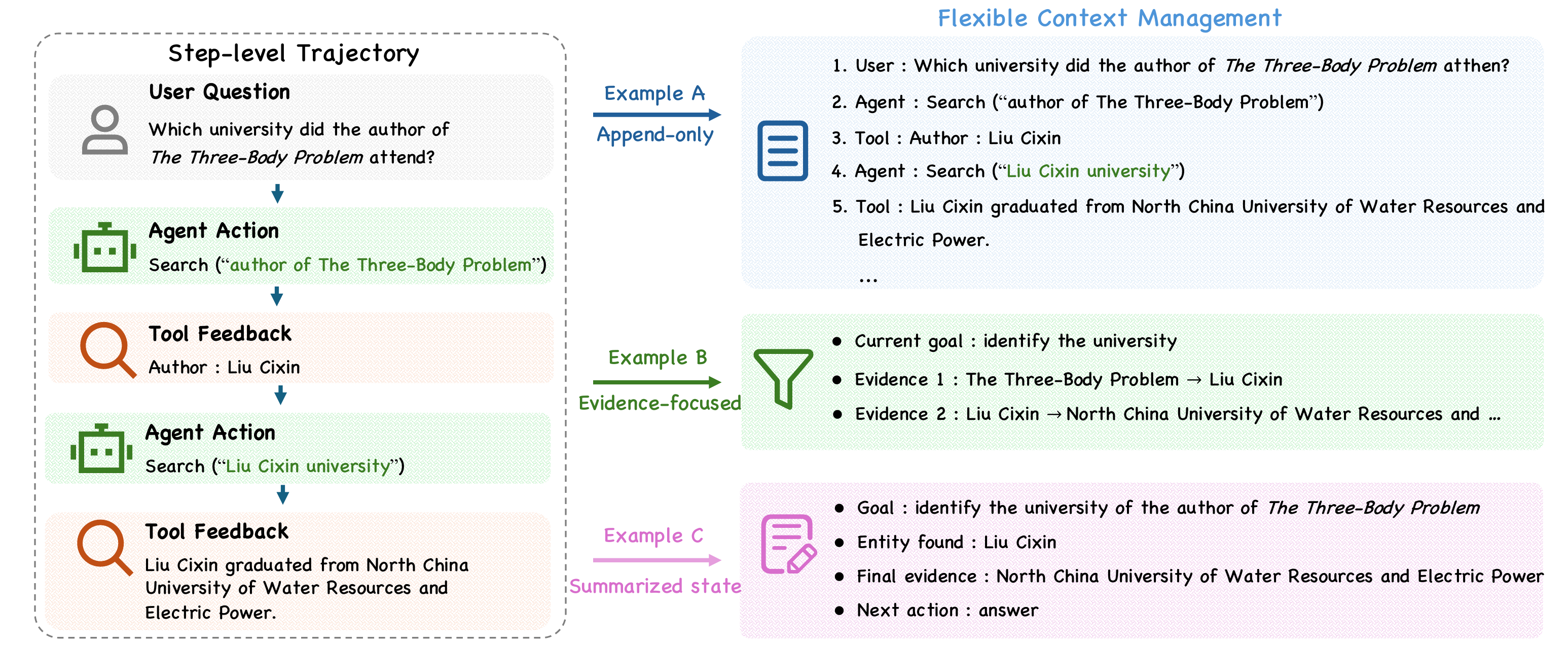}
\caption{A multi-hop question-answering example of flexible context management in Agent-R1. The same step-level replay record can support multiple environment-defined next-observation constructions, including append-only history growth, evidence-focused selection, and summarized task state. }
\label{fig:context-management-case}
\end{figure*}

\subsection{Flexible Context Management}

Flexible context management is the second key design choice in Agent-R1. As illustrated in Figure~\ref{fig:context-management-case}, the next context in a multi-turn agent system can be constructed in multiple ways. This is important because tool outputs may be verbose, intermediate reasoning may be irrelevant, and long interaction histories may exceed the useful context budget for the next decision. Rather than hard-wiring a fixed append-only strategy, Agent-R1 allows the next observation to be defined by an environment-specific context rule:
\begin{equation}
o_{t+1} = \mathcal{C}(z_0, z_1, \ldots, z_t),
\end{equation}
where the visible context is constructed from structured interaction history rather than blind concatenation. This allows prior traces to be preserved, summarized, omitted, or otherwise transformed while remaining in the replay record, and naturally supports more general memory-management strategies.

This flexibility is one of the key reasons step-level representation matters. Once trajectories are stored as structured steps rather than one flat text stream, context construction no longer needs to be identical to raw replay. Agent-R1 can keep an exact replayable record for optimization while allowing the environment to decide what should be exposed to the model at the next step. In this sense, trajectory representation and context management are not independent features, but two sides of the same step-level design choice.

\section{Experiments}
\label{sec:research_path}

We evaluate two questions: whether Agent-R1 transfers across different agent tasks, and whether its context-management interface affects learning quality under a fixed training setup.

\subsection{Experimental Setting}

We instantiate Agent-R1 with Qwen3-4B\cite{yang2025qwen3} on GSM8K\cite{cobbe2021training}, HotpotQA\cite{yang2018hotpotqa}, ALFWorld\cite{shridhar2020alfworld}, and WebShop \cite{yao2022webshop}. 
These tasks span arithmetic reasoning with sandboxed coding, retrieval-based multi-hop question answering, embodied household interaction, and simulated online shopping. The controlled comparisons below focus on GSM8K, where the Agent-R1 environment, tool-based interaction setting, tool format, and reward definition are fixed, so that differences can be attributed more directly to the optimization algorithm or context-management rule. The reward combines answer accuracy with a format component.

\subsection{Different Application Scenarios}

\begin{table*}[t]
\centering
\caption{Main experimental results across representative application scenarios under Agent-R1. The best result in each column is in bold, and the second-best result is underlined.}
\label{tab:main_results}
\footnotesize
\renewcommand{\arraystretch}{1.08}
\setlength{\tabcolsep}{10pt}
\begin{tabular}{@{}lcccccc@{}}
\toprule
Method & GSM8K & HotpotQA & \multicolumn{2}{c}{ALFWorld} & \multicolumn{2}{c}{WebShop} \\
\cmidrule(lr){4-5}\cmidrule(lr){6-7}
 & Acc. (\%) & Acc. (\%) & SR (\%, Seen) & SR (\%, Unseen) & Score (\%) & SR (\%) \\
\midrule
ReAct & 53.1 & 25.8 & 7.14 & 2.98 & 51.58 & 23.8 \\
GRPO & \textbf{83.3} & \textbf{59.4} & \textbf{81.29} & \textbf{74.58} & 65.83 & 44.2 \\
PPO & 78.1 & \underline{56.7} & 76.42 & 72.38 & \textbf{70.18} & \textbf{46.0} \\
Reinforce++ & 78.9 & 52.8 & 73.84 & 69.57 & 63.41 & 41.8 \\
RLOO & \underline{81.6} & 55.2 & \underline{79.08} & \underline{73.46} & \underline{68.02} & \underline{45.1} \\
\bottomrule
\end{tabular}
\end{table*}

\begin{figure*}[t]
\centering
\includegraphics[width=\textwidth]{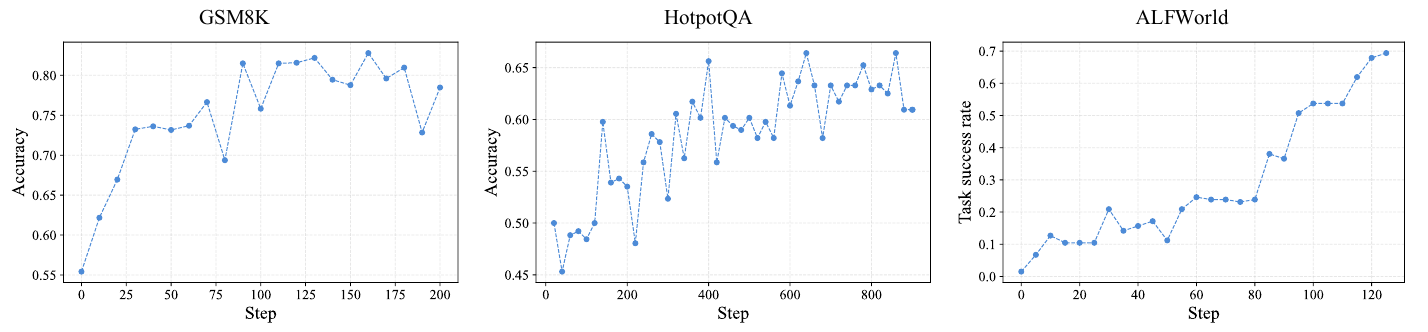}
\caption{Representative training curves of Agent-R1 across multiple application scenarios. We show results on GSM8K, HotpotQA, and ALFWorld as three representative examples to illustrate that the same framework can be instantiated for tool-augmented mathematical reasoning, multi-hop question answering, and interactive decision-making tasks.}
\label{fig:agent_r1_datasets}
\end{figure*}

Table~\ref{tab:main_results} summarizes the main results across representative scenarios. We report one representative task metric for each setting and compare GRPO, PPO, Reinforce++, and RLOO under the same Agent-R1 framework.
All four RL methods outperform the training-free baseline across these diverse settings. At the same time, the best optimizer varies by task: GRPO leads on the arithmetic, retrieval, and embodied settings, while PPO is strongest on the shopping task. This pattern suggests that Agent-R1 is broad enough to support heterogeneous agent environments while still preserving meaningful algorithmic differences.
Figure~\ref{fig:agent_r1_datasets} shows representative training curves on three distinct tasks under GRPO. All three exhibit clear upward trends, indicating that Agent-R1 can support effective learning across heterogeneous agent settings. At the same time, their optimization dynamics differ substantially: GSM8K improves rapidly and stabilizes early at a high level, HotpotQA shows slower and more fluctuating gains, and ALFWorld exhibits a more stage-wise improvement pattern with pronounced late-stage jumps. This suggests that while the same framework transfers across tasks, the underlying learning dynamics remain task-dependent.

\begin{figure*}[t]
\centering
\includegraphics[width=\textwidth]{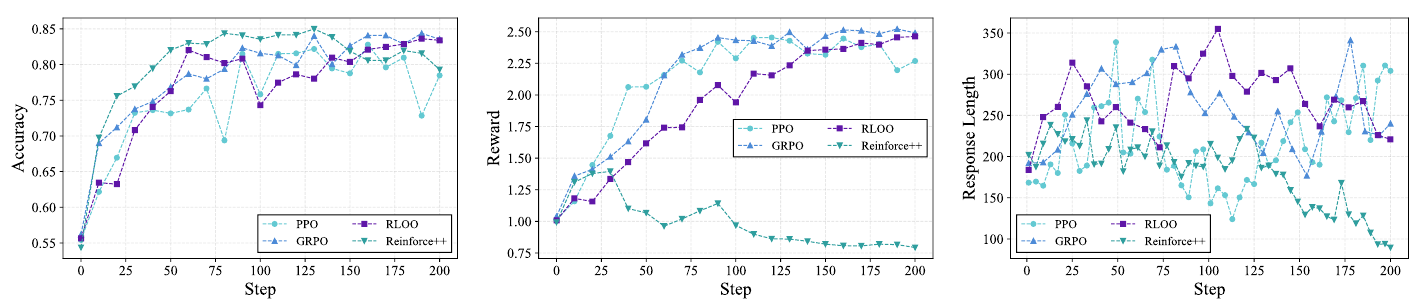}
\caption{Training curves on GSM8K under Agent-R1 with different RL algorithms. We compare PPO, GRPO, Reinforce++, and RLOO under the same environment, tool format, and rollout configuration, and report reward, accuracy, and response length to show how optimizer choice affects both task performance and response behavior.}
\label{fig:gsm8k_curves}
\end{figure*}
\subsection{Different RL Algorithms}

We further compare PPO, GRPO, Reinforce++, and RLOO under the same GSM8K environment to isolate the effect of the optimizer. Figure~\ref{fig:gsm8k_curves} reports reward, accuracy, and response length under matched prompts, tool format, and rollout configuration.
Two patterns are especially notable. First, GRPO and RLOO reach the strongest late-stage accuracy, while PPO remains more volatile. Second, Reinforce++ behaves differently from the other optimizers: although it still achieves relatively high accuracy, its reward remains substantially lower. This discrepancy is consistent with its much shorter responses in the later stage of training. Since the reward in this setting combines answer accuracy with a format-related component, Reinforce++ appears to learn a more conservative policy that can still produce correct answers, but is less effective at maximizing the full training signal. This highlights that, in multi-turn tool-augmented RL, high task accuracy does not necessarily imply high reward, and different optimizers may favor different response strategies. This shows that Agent-R1 does not wash out optimizer-specific behavior; it makes that behavior observable under a common interaction setup.

\subsection{Context-Management Strategies}

\begin{figure*}[t]
\centering
\includegraphics[width=\textwidth]{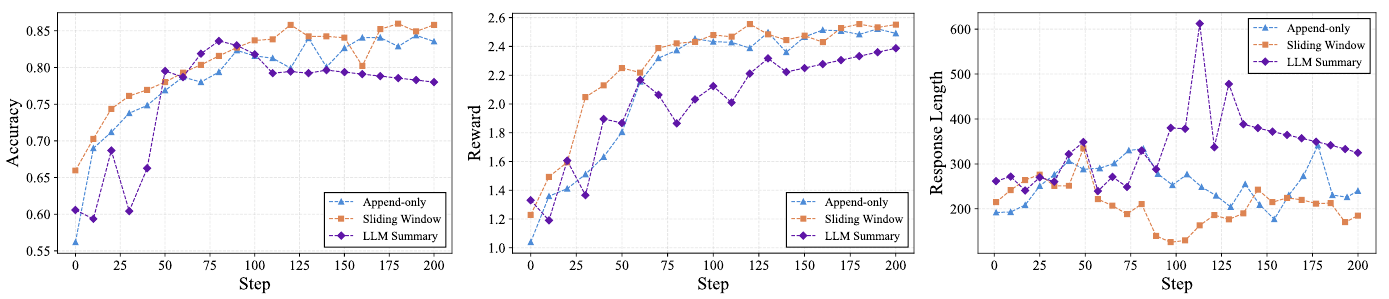}
\caption{Comparison of three context-management strategies on GSM8K under the same GRPO training setup. We compare append-only context, sliding-window context that keeps the original question together with the most recent tool output and model analysis, and LLM-summarized context that compresses the evolving interaction history.}
\label{fig:gsm8k_context_management}
\end{figure*}

To test whether flexible context construction matters in practice, we compare append-only, sliding-window, and LLM-summarized context under the same GRPO setup on GSM8K. Figure~\ref{fig:gsm8k_context_management} shows that sliding-window context performs best, while direct append-only replay is weaker and summary-based context underperforms in this small-model setting.
The result supports the main design claim of Agent-R1: context management is not just a presentation detail. When the framework exposes context construction explicitly, it becomes possible to study how different memory rules affect training under the same rollout and optimizer. In this experiment, preserving only the most relevant recent evidence produces a cleaner learning signal than either unbounded history growth or noisy model-generated summaries. The poor performance of LLM-summarized context in this setting should not be read as a general rejection of summary-based memory; it instead suggests that the quality of the transformation itself becomes part of the training problem. The response-length curves are broadly consistent with this interpretation: sliding-window context keeps responses more controlled without hurting task performance, whereas the other two strategies either retain excessive history or introduce lossy compression.

\section{Future Directions}
\label{sec:experiments}

Agent-R1 is designed as a starting point rather than an endpoint, and several future directions arise from the step-level view itself. A first question is what the next generation of trajectory representation should look like. The current step-level trace already makes context management more flexible and creates a cleaner substrate for step-level credit assignment, but it can still introduce computational redundancy during training because many trajectories share long common prefixes while being optimized separately. Recent systems such as MiniMax Forge have begun to explore prefix-sharing and tree-structured merging to reduce repeated computation across related trajectories \cite{forge2026}. A natural next step is therefore to ask how structured step traces can preserve replay fidelity and context flexibility while also exposing more efficient computation graphs for optimization. A second direction concerns how increasingly complex agent environments should be connected to the training framework. As agents move into settings with richer tools, branching workflows, persistent state, and delayed feedback, the framework should make environment integration as non-intrusive as possible without drifting too far from the on-policy interaction that the optimizer assumes \cite{luo2025agentlightning,fu2025areal}. Balancing these two goals is difficult: tighter integration often improves faithfulness, while looser integration often improves usability. A third direction is how to obtain higher-quality RL data for agent training. In multi-turn settings, data quality is determined not only by final success or failure, but also by whether trajectories contain informative intermediate decisions, meaningful tool interactions, and useful exploration patterns \cite{wang2025ragen,jin2025search}. This suggests that future progress may depend as much on better data generation, filtering, and curriculum design as on the optimizer itself. In this sense, the long-term challenge is not only to make agent training easier to run, but to identify the right data and abstraction for learning robust multi-turn agent behavior.

\section{Limitation Discussion}

Although Agent-R1 provides a unified and modular framework for agentic reinforcement learning, it still has several limitations. First, the step-level trajectory representation may bring extra training cost. When an agent uses append-only history without explicit context management, different training samples may share long prefixes. Since these step-level records are trained separately, the same prefix can be computed many times, causing redundant computation. This issue becomes more serious for long-horizon agents with many interaction steps. Future work may reduce this cost through prefix sharing, KV-cache reuse, or tree-structured trajectory merging, while keeping the replay fidelity and context flexibility of step-level traces.

Second, Agent-R1 has not fully explored asynchronous rollout-training execution. Although rollout workers and the training backend are decoupled, the current system still requires collected trajectories to be aligned with optimization updates. This design is practical for simple or moderately long tasks, but it can be inefficient for complex agents with slow tools, branching workflows, persistent environments, or delayed feedback. In these cases, rollout may become the main bottleneck, making training less efficient than fully asynchronous systems. Improving asynchronous scheduling and scalable rollout-training coordination is an important direction for future work.
\section{Related Work}
\label{sec:related_work}

This section situates Agent-R1 within the emerging literature on agentic RL, with emphasis on two closely related questions: how multi-turn agent behavior is optimized, and how training frameworks organize that optimization in practice across different agent settings.

\subsection{Agentic RL Algorithms}

Agentic RL algorithms inherit their basic optimization toolbox from earlier RL post-training work such as PPO \cite{schulman2017ppo} and GRPO-style training \cite{shao2024deepseekmath}, but they must adapt these ideas to multi-turn interaction, sparse feedback, and environment-coupled behavior. Early work such as Search-R1 \cite{jin2025search} and RAGEN \cite{wang2025ragen} already made these difficulties visible by showing that response-level rewards are often too weak to supervise long interaction chains. A natural trend that followed was to move from flat sequence optimization toward turn-aware and step-aware credit assignment. This transition is also related to broader sequence-level directions such as DAPO \cite{yu2025dapo} and GSPO \cite{zheng2025group}, which make the optimization unit less token-local even outside fully agentic settings. Many later methods can then be understood as extensions of the PPO and GRPO families in a more explicitly agentic direction. On the PPO side, Turn-PPO \cite{li2026turn} reformulates multi-turn training around turn-level advantage estimation, StepPO \cite{wang2026steppo} pushes this idea further by aligning credit assignment with agent steps rather than individual tokens, and PaperScout \cite{pan2026paperscout} emphasizes process-aware sequence-level optimization for literature-search agents. On the GRPO side, Tree Search for LLM Agent Reinforcement Learning \cite{ji2025tree} and GiGPO \cite{wang2025gigpo} both move beyond flat response-level grouping and attempt to expose finer-grained structure inside multi-step agent rollouts. In parallel, another line of work focuses less on changing the base optimizer itself and more on strengthening the supervision signal around it. AgentPRM \cite{xi2026agentprm} studies step-wise process reward modeling for agent trajectories, and SWEET-RL \cite{zhou2025sweet} explores the use of privileged training-time information to improve supervision over multi-turn reasoning behavior. Taken together, these methods suggest a clear trend: agentic RL is moving away from treating the entire rollout as a single undifferentiated response, and toward optimization schemes that respect the internal structure of interaction.

\subsection{Agentic RL Training Frameworks}

In parallel with algorithmic advances, agentic RL training frameworks have also become a distinct and rapidly evolving research topic. One line of development starts from general RL post-training substrates and gradually extends them toward agent workloads. veRL \cite{sheng2025hybridflow} provides a strong foundation for distributed RL post-training, while slime \cite{slime2025} emphasizes high-performance scaling and flexible generation workflows. A second line focuses on how to connect already-built agents to RL with minimal friction. Agent Lightning \cite{luo2025agentlightning} foregrounds decoupled agent-training architecture and broad compatibility with existing agents, while rLLM \cite{rllm2026} emphasizes low-intrusion integration with external agent frameworks. A third line places more weight on execution topology and system scalability. AReaL \cite{fu2025areal} focuses on asynchronous rollout-training execution, MiniMax Forge \cite{forge2026} highlights efficient large-scale agent RL workloads, and Claw-R1 \cite{clawr1repo} extends this design space toward a more deployment-oriented framework with stronger emphasis on complex runtime integration. The broader trend across these systems is a movement from generic RL infrastructure toward increasingly agent-native training abstractions. Agent-R1 is positioned in this latter direction. Its main emphasis is the training abstraction itself: step-level MDP, structured step-native traces, flexible context management, and interfaces that let multiple optimization algorithms share the same multi-turn substrate \cite{agentr1repo}. In this sense, Agent-R1 is less about one specific optimizer and more about making agentic RL trainable through a coherent framework design.

\section{Conclusion}
\label{sec:conclusion}

In this paper, we first revisited how LLM training extends into RL training, and why the transition from single-turn post-training to agentic RL brings new challenges in environment adaptation, rollout structure, trajectory representation, and context construction. We then introduced Agent-R1 as a unified and modular framework for multi-turn RL training of LLM agents, with emphasis on step-level MDP abstraction, structured step-level trajectories, flexible context management, unified interfaces across workflows and infrastructure, and compatibility with multiple optimization algorithms such as PPO and GRPO. We further showed, through a controlled GSM8K case study and the broader framework design throughout the paper, how these ideas make it possible to support different algorithmic instantiations on top of a common multi-turn RL substrate while keeping the training process analyzable, extensible, and reusable. As LLM agents continue to grow in capability and complexity, we hope Agent-R1 can provide the community with a principled, modular, and easy-to-use foundation for building, studying, and extending future agentic RL systems.

\bibliographystyle{plainnat}
\bibliography{main}

\end{document}